\documentclass{article}

\usepackage{spconf,amsmath,amssymb,graphicx}
\usepackage{bm}
\usepackage{tikz}
\usepackage{booktabs}
\usepackage{multirow}
\usepackage{algorithm,algpseudocode}
\usepackage[shortlabels]{enumitem}
\usepackage{hyperref}
\hypersetup{hidelinks}

\usetikzlibrary{arrows.meta,fit,positioning}

\newcommand{\given}{\,|\,}
\newcommand{\Normal}{\mathcal{N}}
\newcommand{\E}{\mathbb{E}}

\title{Direct Conditional Transition Sampling for Diffusion Inverse Problems}

\name{Qi Yu, Hanlin Wu, Xiaohui Sun}
\address{School of Information Science and Technology, Beijing Foreign Studies University}

\begin{document}

\maketitle

\begin{abstract}
Training-free diffusion inverse solvers typically choose between local measurement guidance and costly clean-space posterior updates.  Independent posterior refresh can improve global correction by sampling a clean conditional and re-noising it, but its practical realization requires probability-flow ODE integration and clean-space Markov chain Monte Carlo (MCMC).  We propose Direct Conditional Transition Sampling (DCTS), a direct stochastic-flow approximation to the same ideal refresh target.  Rather than explicitly drawing a clean sample, DCTS estimates the measurement-conditioned clean mean along a short inner path and transports Gaussian source noise directly to the next noisy state. A denoiser-compatible sufficient statistic and a covariance-scaled operator update enable this conditional-mean estimation. Experiments on four inverse problems demonstrate that DCTS achieves competitive reconstruction quality with up to $16.8\times$ speedups over competing methods.
\end{abstract}


\section{Introduction}

The goal of inverse problems is to recover an unknown signal $z\in\mathbb R^d$ from an observation $\bm y =\mathcal A(\bm z)+\bm n$, where $\mathcal A:\mathbb R^d\rightarrow\mathbb R^m$ is a known forward operator and $\bm n\sim\Normal(\bm 0,\bm\Sigma_y)$ is additive Gaussian noise.  The problem is inherently ill-posed, as the potential nonlinearity and information loss of $\mathcal{A}$ can lead to multiple plausible solutions for a single observation. 
A Bayesian formulation characterizes this uncertainty using the posterior distribution
\begin{equation}
p(\bm z\given\bm y)\propto
p(\bm y\given\bm z)p_{\rm data}(\bm z),
\end{equation}
where $p_{\rm data}$ is the prior distribution and $p(\bm y\given\bm z)$ is the likelihood induced by the measurement model. 

\begin{figure}[t]
    \includegraphics[width=0.9\linewidth]{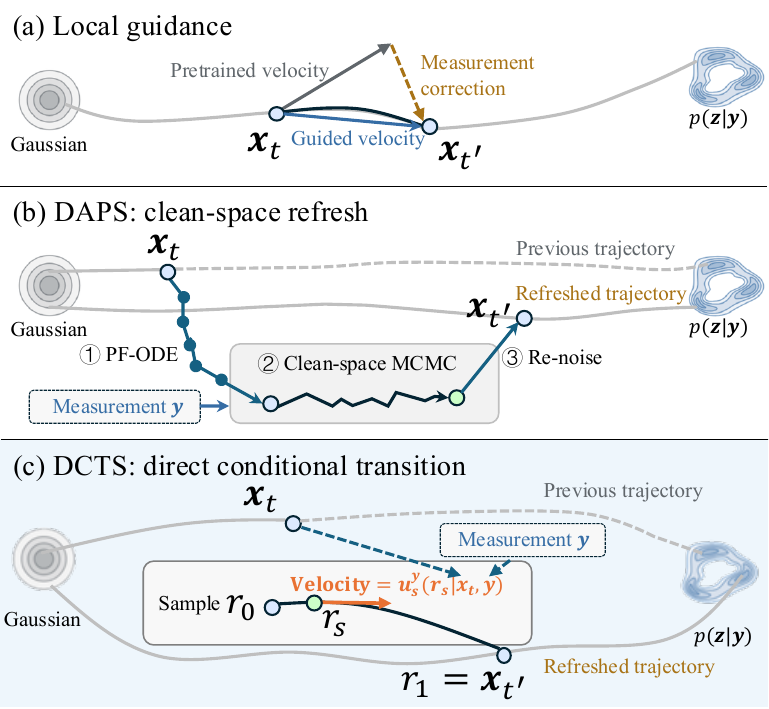}
    \caption{Comparison of three conditional transitions: (a) local guidance; (b) DAPS clean-space refresh; and (c) our proposed DCTS, which directly transports noise to the next state.}
    \label{fig:transition-comparison}
    % \vspace{-5pt}
\end{figure}

Diffusion models~\cite{ho2020ddpm,song2021score} provide an expressive learned prior and can therefore solve various inverse problems with one pretrained denoiser, without retraining a conditional model for each operator.  The forward diffusion process generates a noisy state $X_t$ from a clean signal $Z$ through Gaussian perturbation:
\begin{equation*}
X_t=\alpha_t Z+\sigma_t\bm\epsilon_t,
\qquad \bm\epsilon_t\sim\Normal(\bm0,\bm I),
\end{equation*}
where $\alpha_t$ and $\sigma_t$ are known schedule coefficients. Equivalently, the forward diffusion kernel is
\begin{equation}
q_t(\bm x\given\bm z)
:=\Normal(\bm x;\alpha_t\bm z,\sigma_t^2\bm I).
\label{eq:probability-path}
\end{equation}
We adopt the variance-exploding (VE) parameterization~\cite{song2021score}, for which $\alpha_t\equiv1$. Using score estimates from a pretrained denoiser, we can numerically solve the reverse-time stochastic differential equation (SDE) to generate samples from the learned data prior. For inverse problems, our goal is to use this denoiser to sample from the posterior $p(\bm z\given\bm y)$, which requires incorporating the measurement $\bm y$ into the sampling process.

Existing training-free solvers employ two broad strategies for posterior sampling: local guidance and clean-space updates. Guidance methods modify each reverse diffusion step using a tractable approximation to the noisy likelihood score. DPS~\cite{chung2023dps} differentiates a measurement residual evaluated at a denoised estimate, whereas DDRM~\cite{kawar2022ddrm}, PiGDM~\cite{song2023pigdm}, and CoDPS~\cite{yismaw2025codps} exploit linear-operator structure or a Gaussian approximation. TMPD~\cite{boys2024tmpd} and optimal posterior covariance methods~\cite{peng2024covariance} further incorporate denoiser uncertainty into the measurement update. However, small updates between consecutive noisy states can make it difficult to correct global errors from earlier sampling steps, particularly in nonlinear inverse problems~\cite{zhang2025daps}.

A second class incorporates measurements through explicit updates in clean space, then generates the next noisy state. DDNM~\cite{wang2023ddnm} enforces data consistency in denoised estimates, while DiffPIR~\cite{zhu2023diffpir} solves a clean-space data subproblem between denoising steps.  DAPS~\cite{zhang2025daps} refreshes the noisy state through clean-space sampling and independent re-noising, allowing transitions beyond local trajectory corrections. Each refresh involves three steps: (i) integrating a probability-flow ODE from $\bm x_t$ to obtain a clean estimate; (ii) initializing MCMC from this estimate to approximately sample $Z$ from $p(\bm z\given\bm x_t,\bm y)$; and (iii) re-noising via $X_{t'}=Z+\sigma_{t'}\bm\xi$, where $\bm\xi\sim\Normal(\bm0,\bm I)$ is fresh Gaussian noise. The cost of repeated ODE integration and MCMC sampling motivates the following question:

\textit{Can we realize the same posterior refresh through a direct transition from $\bm x_t$ to $\bm x_{t'}$, without explicit clean-space sampling and re-noising?}

We answer this question with \emph{Direct Conditional Transition Sampling} (DCTS), a training-free method that replaces explicit clean-space sampling and re-noising with a direct transition based on Flow Matching~\cite{lipman2023flow}. Along this flow, we combine the current noisy state $\bm x_t$ and the evolving flow state into a single Gaussian-corrupted state compatible with the pretrained denoiser. The denoiser provides a clean-signal estimate, which we refine using the measurement $\bm y$ through a Gaussian update tailored to the forward operator. This refined estimate determines the flow velocity that transports fresh Gaussian noise directly to the next noisy state $\bm x_{t'}$.
As illustrated in Fig.~\ref{fig:transition-comparison}, the resulting transition retains the same posterior-refresh target while avoiding explicit clean-space sampling, MCMC refinement, and re-noising.
Our contributions are threefold:
\begin{enumerate}[1),nosep]
    \item We formulate independent posterior refresh as a direct stochastic transport, enabling global corrections beyond the constraints of local trajectory updates.
    \item We develop a training-free method for estimating transport velocities conditioned on measurements using a pretrained denoiser.
    \item Experiments on diverse inverse problems demonstrate that DCTS achieves competitive reconstruction quality with substantially reduced computational cost.
\end{enumerate}

\section{Method}
\label{sec:method}

\subsection{Problem Formulation}

Conditioned on an observation $\bm y$, the forward diffusion kernel in Eq.~\eqref{eq:probability-path} defines the noisy posterior marginal
\begin{equation}
\pi_t(\bm x\given\bm y)
:=\int q_t(\bm x\given\bm z)\,p(\bm z\given\bm y)\,\mathrm{d}\bm z.
\label{eq:target-marginal}
\end{equation}
We aim to realize the independent posterior-refresh kernel:
\begin{equation}
K^\star_{t\rightarrow t'}(\bm x'\given\bm x,\bm y)
=\int q_{t'}(\bm x'\given\bm z)
p(\bm z\given\bm x,\bm y)\,\mathrm{d}\bm z.
\label{eq:refresh-kernel}
\end{equation}
This kernel maps the noisy posterior marginal $\pi_t$ to $\pi_{t'}$: if $X_t\sim\pi_t(\cdot\given\bm y)$, sampling $Z\sim p(\cdot\given X_t,\bm y)$ yields the clean posterior marginal $p(\bm z\given\bm y)$, and re-noising through $q_{t'}$ yields $\pi_{t'}(\cdot\given\bm y)$. The posterior refresh kernel defined in Eq.~\eqref{eq:refresh-kernel} is the same as that of DAPS~\cite{zhang2025daps}, but DCTS realizes it differently. DAPS explicitly approximates the clean draw from $p(\cdot\given\bm x_t,\bm y)$ and then re-noises it. In contrast, we define an inner conditional flow for each outer step $t\to t'$. Conditioned on $\bm x_t$ and $\bm y$, we sample $\bm x_{t'}$ by integrating the flow ODE from fresh Gaussian noise, bypassing explicit clean-space sampling and re-noising. We next present a training-free construction of the flow's velocity field using a pretrained denoiser.

\subsection{Stochastic Inner Path}

We build on the GLASS~\cite{holderrieth2026glass} construction of Gaussian transitions, adapting it to inverse problems by using the measurements and the forward operator to refine the denoiser's clean-signal estimate. Let $s\in[0,1]$ denote an inner coordinate, distinct from the outer diffusion time. We linearly interpolate between a Gaussian source and the refresh endpoint:
\begin{equation}
R_s=(1-s)\sigma_{\rm src}\bm\eta+s(Z+\sigma_{t'}\bm\eta),\quad \bm\eta\sim\Normal(\bm0,\bm I).
\label{eq:inner-path}
\end{equation}
Here $\sigma_{\rm src}>0$ is an auxiliary source scale and $\bm\eta$ is independent of $(Z,X_t,\bm y)$.  Equivalently, $R_s=sZ+\bar\sigma_s\bm\eta$, where $\bar\sigma_s=(1-s)\sigma_{\rm src}+s\sigma_{t'}$.  The source $R_0=\sigma_{\rm src}\bm\eta$ is easy to sample, while $R_1=Z+\sigma_{t'}\bm\eta$ has the endpoint law in Eq.~\eqref{eq:refresh-kernel}.  By the marginal velocity identity of Flow Matching~\cite{lipman2023flow}, the marginal velocity at $\bm r_s$ is the conditional expectation of the sample-wise path velocity given the current state and conditioning variables.  For the linear path in Eq.~\eqref{eq:inner-path}, this expectation depends on the clean endpoint only through the conditional clean mean
\begin{equation}
\bm m_s^\star:=\E[Z\given R_s=\bm r_s,X_t=\bm x_t,\bm y].
\label{eq:conditional-clean-mean}
\end{equation}
Therefore, estimating $\bm m_s^\star$ along the inner path is sufficient to construct the velocity field.

\subsection{Measurement-Conditioned Clean Mean}

We estimate $\bm m_s^\star$ in two stages. First, before incorporating $\bm y$, we estimate the prior clean mean $\E[Z\given X_t=\bm x_t,R_s=\bm r_s]$. This mean depends on two noisy states, whereas a pretrained denoiser accepts only a single Gaussian-corrupted state and its noise level. Since $X_t$ and $R_s$ are conditionally independent given $Z$, we follow the derivation in~\cite{holderrieth2026glass} to combine these two Gaussian observations into a single sufficient statistic for $Z$ that preserves this conditional mean:
\begin{equation}
\widetilde X_s:=v_s\bigg(\frac{X_t}{\sigma_t^2}+\frac{sR_s}{\bar\sigma_s^2}\bigg)=Z+\widetilde\sigma_s\widetilde{\bm\epsilon}_s,
\label{eq:sufficient-statistic}
\end{equation}
where $v_s=(\sigma_t^{-2}+s^2/\bar\sigma_s^2)^{-1}$, $\widetilde\sigma_s=\sqrt{v_s}$, and $\widetilde{\bm\epsilon}_s\sim\Normal(\bm0,\bm I)$. The first equality defines $\widetilde X_s$ as a precision-weighted average of the two states, and the second shows that it remains a Gaussian-corrupted version of $Z$.  At $s=0$, $R_0$ contains no information about $Z$, so $\widetilde X_0=X_t$ and $\widetilde\sigma_0=\sigma_t$; as $s$ increases, $\widetilde X_s$ progressively incorporates information from the evolving inner state.  For the observed states $(\bm x_t,\bm r_s)$, its realization $\widetilde{\bm x}_s=v_s(\bm x_t/\sigma_t^2+s\bm r_s/\bar\sigma_s^2)$ and noise level $\widetilde\sigma_s$ therefore form a valid denoiser input.  Moreover, the sufficiency of $\widetilde X_s$ for $Z$ ensures that 
$
\E[Z\given X_t=\bm x_t,R_s=\bm r_s]=\E[Z\given\widetilde X_s=\widetilde{\bm x}_s].
$
By Tweedie's formula~\cite{efron2011tweedie}, one evaluation of the pretrained denoiser $D$ on this Gaussian-corrupted statistic estimates the same clean conditional mean:
\begin{equation}
\bm m_s=D(\widetilde{\bm x}_s,\widetilde\sigma_s)
\approx\E[Z\given X_t=\bm x_t,R_s=\bm r_s].
\label{eq:prior-mean}
\end{equation}

In the second stage, we incorporate $\bm y$ using a local Gaussian approximation~\cite{song2023pigdm} with covariance scale $\kappa>0$:
\begin{equation}
p(\bm z\given\bm x_t,\bm r_s)\approx
\Normal(\bm z;\bm m_s,c_s\bm I),\quad
c_s=\kappa v_s,
\label{eq:local-gaussian}
\end{equation}

\textbf{Linear operators.} Following Gaussian conditioning approaches~\cite{song2023pigdm,yismaw2025codps}, for $\mathcal A(\bm z)=\bm A\bm z$, Gaussian conditioning corrects $\bm m_s$ to
\begin{equation}
\bm m_s^{\bm y}=\bm m_s+c_s\bm A^\top
(c_s\bm A\bm A^\top+\bm\Sigma_y)^{-1}
(\bm y-\bm A\bm m_s).
\label{eq:linear-moment-update}
\end{equation}
The inverse balances this local covariance against measurement noise, making Eq.~\eqref{eq:linear-moment-update} the exact posterior mean under approximation~\eqref{eq:local-gaussian}. Eq.~\eqref{eq:linear-moment-update} is evaluated using operator-specific closed-form or iterative solvers without forming the inverse explicitly, with implementation details in Sec.~\ref{sec:experimental-setup}.

\textbf{Nonlinear operators.} For a differentiable nonlinear operator, let $\bm J_s=\nabla\mathcal A(\bm m_s)$ denote its Jacobian at the current mean.  Substituting its local linearization into the Gaussian update gives the Gauss--Newton moment
\begin{equation}
\bm m_s^{\bm y}=\bm m_s+c_s\bm J_s^\top
(c_s\bm J_s\bm J_s^\top+\bm\Sigma_y)^{-1}
[\bm y-\mathcal A(\bm m_s)].
\label{eq:nonlinear-moment-update}
\end{equation}
We evaluate Eq.~\eqref{eq:nonlinear-moment-update} using Jacobian-vector products without explicitly forming $\bm J_s$; implementation details are provided in Sec.~\ref{sec:experimental-setup}.  Unlike the linear update, the nonlinear update is only locally valid and can be inaccurate for strongly multimodal conditionals.

\subsection{Direct Transition Velocity}

Having estimated the measurement-conditioned clean moment, we now convert it into the velocity that transports the Gaussian source to the refresh endpoint.  Differentiating the inner path in Eq.~\eqref{eq:inner-path} and substituting $\bm\eta=(R_s-sZ)/\bar\sigma_s$ gives the sample-wise velocity $\dot R_s=b_sR_s+(1-sb_s)Z$, where $b_s=(\sigma_{t'}-\sigma_{\rm src})/\bar\sigma_s$ for $s\in[0,1)$.  Taking its conditional expectation and replacing the unknown clean conditional mean by $\bm m_s^{\bm y}$ gives the practical velocity field
\begin{equation}
\bm u_s^{\bm y}=b_s\bm r_s+(1-sb_s)\bm m_s^{\bm y}.
\label{eq:conditional-velocity}
\end{equation}
The first term changes the innovation scale from $\sigma_{\rm src}$ to $\sigma_{t'}$, while the second transports the measurement-conditioned clean content.  If $\bm m_s^{\bm y}=\bm m_s^\star$, this is the exact marginal velocity for the inner path, and exact integration recovers the endpoint law in Eq.~\eqref{eq:refresh-kernel} without explicitly drawing $Z$.  In practice, DCTS recomputes the approximate moment $\bm m_s^{\bm y}$ at each numerical inner state.

For each outer transition $t\rightarrow t'$, we first draw a fresh source $\bm r_0=\sigma_{\rm src}\bm\eta$ and set $\Delta s=1/M$.  At each $s=j/M$, we evaluate the conditional moment and velocity at the current $\bm r_s$, then apply the Euler update
\begin{equation}
\bm r_{s+\Delta s}=\bm r_s+\Delta s\,\bm u_s^{\bm y}.
\label{eq:euler-update}
\end{equation}
After $M$ steps, we set $\bm x_{t'}=\bm r_1$.  Each inner step requires one denoiser forward and one measurement-conditioned moment update, so $L$ outer transitions use $LM$ of each, with no denoiser backward pass or variable-length clean-space chain.  Algorithm~\ref{alg:dcts} summarizes the complete sampler.

\begin{algorithm}[t]
\caption{Direct Conditional Transition Sampling}
\label{alg:dcts}
\begin{algorithmic}[1]
\Require Denoiser $D$, measurement $(\bm y,\mathcal A,\bm\Sigma_y)$, schedule
$\sigma_0>\cdots>\sigma_L=0$, inner steps $M$, $\sigma_{\rm src}$, and $\kappa$
\State Sample $\bm x^{(0)}\sim\Normal(\bm0,\sigma_0^2\bm I)$
\For{$k=0,\ldots,L-1$}
    \State Set $(\bm x_t,\sigma_t,\sigma_{t'})=(\bm x^{(k)},\sigma_k,\sigma_{k+1})$
    \State Sample $\bm r_0=\sigma_{\rm src}\bm\eta$,
    $\bm\eta\sim\Normal(\bm0,\bm I)$
    \For{$j=0,\ldots,M-1$}
        \State Set $s=j/M$ and form
        $(\widetilde{\bm x}_s,\widetilde\sigma_s)$ by
        Eq.~\eqref{eq:sufficient-statistic}
        \State $\bm m_s\gets D(\widetilde{\bm x}_s,\widetilde\sigma_s)$
        \State Compute $\bm m_s^{\bm y}$ by
        Eq.~\eqref{eq:linear-moment-update} or~\eqref{eq:nonlinear-moment-update}
        \State Update $\bm r_s$ by
        Eqs.~\eqref{eq:conditional-velocity}--\eqref{eq:euler-update}
    \EndFor
    \State $\bm x^{(k+1)}\gets \bm r_1$
\EndFor
\State \Return $\bm x^{(L)}$
\end{algorithmic}
\end{algorithm}

\section{Experiments}
\label{sec:experiments}

\subsection{Experimental Setup}
\label{sec:experimental-setup}

\textbf{Dataset and metrics.} Following DAPS~\cite{zhang2025daps}, we use the first 100 test images from the FFHQ-256 dataset~\cite{karras2019stylegan} to evaluate the effectiveness of DCTS. We report PSNR, SSIM, LPIPS, and ArcFace identity similarity~\cite{deng2019arcface}. LPIPS measures the perceptual difference between the reconstruction and the reference image (lower is better), while ArcFace assesses facial identity preservation (higher is better).

\textbf{Inverse problems.} We consider four inverse problems: super-resolution (SR), phase retrieval, motion deblurring, and grayscale colorization, all with Gaussian measurement noise $\sigma_y=0.05$. For SR, we adopt the antialiased $4\times$ downsampling operator from DAPS~\cite{zhang2025daps}. Phase retrieval is a nonlinear inverse problem of reconstructing an image from its Fourier magnitudes without phase information; we use an oversampling rate of 2.0. For motion deblurring, we convolve images with a $61\times61$ motion blur kernel using reflection padding. Colorization uses a rank-one linear mapping from each RGB pixel to a scalar grayscale measurement.

\textbf{Implementation details.} We adopt the pretrained DDPM from DAPS~\cite{zhang2025daps} as our denoiser. We set the source scale to $\sigma_{\rm src}=1$ and use $L=150$ outer steps and $M=2$ inner steps. For phase retrieval, we increase
$M$ to 8 to improve reconstruction quality. The outer noise levels use the EDM discretization~\cite{karras2022edm} with polynomial exponent 7, $\sigma_{\max}=100$, $\sigma_{\min}=0.1$, and a final noise level of zero. The local covariance is $c_s\bm I$ with $c_s=\kappa v_s$, where $\kappa$ is $1.0$ for SR, $0.5$ for phase retrieval and motion blur, and $0.25$ for colorization.

To evaluate Eq.~\eqref{eq:linear-moment-update}, we exploit the separable structure of the downsampling operator for SR and use pixelwise closed-form updates for colorization. For motion deblurring, we approximately solve the corresponding linear
system using four conjugate-gradient iterations. For phase retrieval, we evaluate Eq.~\eqref{eq:nonlinear-moment-update} using analytic Jacobian products and four conjugate-gradient iterations.

\subsection{Main Results}

\begin{figure}[t]
\centering
\includegraphics[width=\columnwidth]{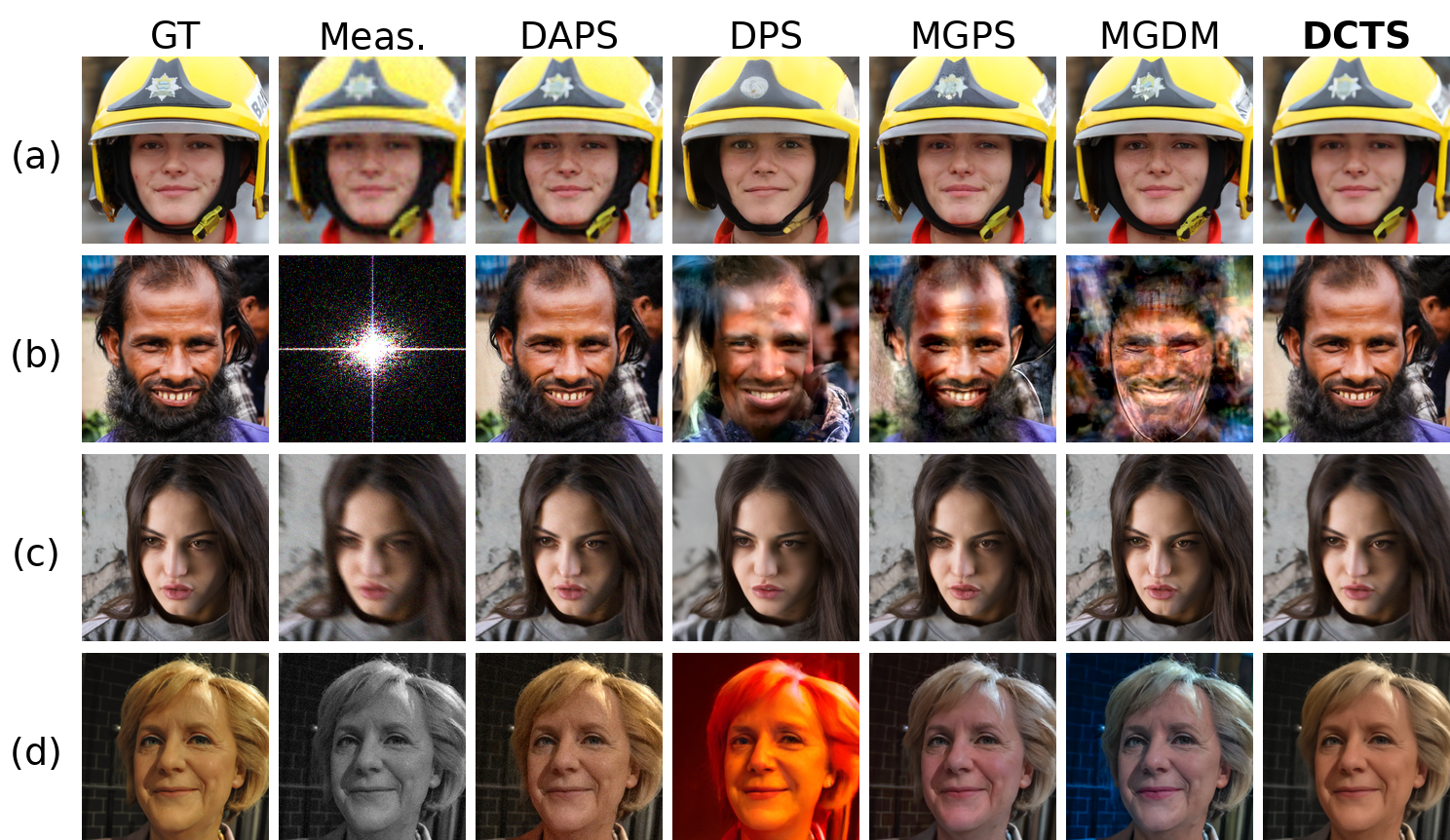}
\caption{Visual comparison for (a) SR $\times$ 4; (b) phase retrieval; (c) deblurring; (d) colorization.}
\label{fig:qualitative}
\end{figure}

\begin{table}[t]
\caption{Quantitative results on FFHQ-256. Best and second-best results are shown in bold and underline, respectively.}
\label{tab:main-results}
\centering
\scriptsize
\setlength{\tabcolsep}{1.2pt}
\begin{tabular}{llccccc}
\toprule
Task & Method & PSNR (dB) $\uparrow$ & SSIM $\uparrow$ & LPIPS $\downarrow$ & ArcFace $\uparrow$ & Time (s) $\downarrow$ \\
\midrule
\multirow{5}{*}{SR $\times$4} & DAPS~\cite{zhang2025daps} & \underline{29.360} & 0.783 & 0.192 & \underline{0.926} & 94.74 \\
 & DPS~\cite{chung2023dps} & 24.663 & 0.705 & 0.259 & 0.619 & \underline{53.02} \\
 & MGPS~\cite{moufad2024mgps} & 28.189 & 0.791 & 0.163 & 0.879 & 81.20 \\
 & MGDM~\cite{janati2025mixture} & 28.526 & \underline{0.807} & \textbf{0.151} & 0.900 & 82.10 \\
 & DCTS & \textbf{30.552} & \textbf{0.861} & \underline{0.160} & \textbf{0.928} & \textbf{5.65} \\
\midrule
\multirow{5}{*}{Phase retrieval} & DAPS~\cite{zhang2025daps} & \textbf{29.705} & 0.771 & 0.182 & \textbf{0.918} & 117.30 \\
 & DPS~\cite{chung2023dps} & 19.065 & 0.550 & 0.394 & 0.439 & \underline{53.78} \\
 & MGPS~\cite{moufad2024mgps} & 28.439 & \underline{0.806} & \textbf{0.165} & 0.874 & 81.61 \\
 & MGDM~\cite{janati2025mixture} & 28.665 & 0.801 & 0.170 & 0.859 & 83.55 \\
 & DCTS$_{(M=8)}$ & \underline{29.696} & \textbf{0.812} & \underline{0.170} & \underline{0.908} & \textbf{25.17} \\
\midrule
\multirow{5}{*}{Deblurring} & DAPS~\cite{zhang2025daps} & \underline{31.651} & 0.836 & 0.137 & \textbf{0.971} & 66.26 \\
 & DPS~\cite{chung2023dps} & 27.285 & 0.773 & 0.183 & 0.842 & \underline{53.02} \\
 & MGPS~\cite{moufad2024mgps} & 30.577 & 0.854 & \underline{0.115} & 0.953 & 81.38 \\
 & MGDM~\cite{janati2025mixture} & 31.048 & \underline{0.860} & \textbf{0.103} & 0.962 & 81.48 \\
 & DCTS & \textbf{32.678} & \textbf{0.886} & 0.128 & \underline{0.969} & \textbf{7.41} \\
\midrule
\multirow{5}{*}{Colorization} & DAPS~\cite{zhang2025daps} & \underline{24.055} & 0.763 & 0.310 & \textbf{0.974} & 59.80 \\
 & DPS~\cite{chung2023dps} & 11.700 & 0.526 & 0.594 & 0.761 & \underline{53.37} \\
 & MGPS~\cite{moufad2024mgps} & 23.001 & \underline{0.877} & \textbf{0.217} & 0.964 & 80.22 \\
 & MGDM~\cite{janati2025mixture} & 22.051 & 0.865 & \underline{0.226} & 0.962 & 81.36 \\
 & DCTS & \textbf{24.412} & \textbf{0.892} & 0.235 & \underline{0.967} & \textbf{5.43} \\
\bottomrule
\end{tabular}
\end{table}
We compare with DAPS~\cite{zhang2025daps}, DPS~\cite{chung2023dps}, MGPS~\cite{moufad2024mgps}, and MGDM~\cite{janati2025mixture}, using their official default sampling hyperparameters. Runtime is measured on an RTX 4090 GPU with a batch size of 1. Table~\ref{tab:main-results} shows that DCTS attains the highest PSNR on three tasks, top-two mean ArcFace similarity on all four, and the lowest per-trajectory runtime. These results demonstrate that DCTS achieves competitive reconstruction quality while reducing runtime by avoiding repeated clean-space sampling. Fig.~\ref{fig:qualitative} shows that DCTS better preserves facial structure and details while reducing artifacts.

\subsection{Ablation Study}
\label{sec:transition-ablation}

We conduct an ablation study to evaluate the proposed transition design and the effect of multiple inner updates. All variants use $L=150$ outer steps, the same pretrained denoiser, and the same measurement-conditioning rule. The DDIM~\cite{song2021ddim} variants replace each DCTS transition with eight DDIM substeps starting from the current noisy state. Here, $\eta$ controls the stochasticity of DDIM: $\eta=0$ yields deterministic updates, while $\eta=1$ injects fresh Gaussian noise at each substep. We also evaluate point refresh, which generates the next noisy state by adding fresh Gaussian noise to a single measurement-conditioned clean estimate at each outer step. This corresponds to a single Euler step ($M=1$), whereas full DCTS recomputes the estimate along eight inner steps ($M=8$). Point refresh therefore uses one eighth as many denoiser calls as full DCTS. Table~\ref{tab:transition-ablation} reports the phase-retrieval results on the first 100 FFHQ-256 test images. Full DCTS achieves the best mean values on all four metrics.

\begin{table}[t]
\caption{Ablation study on the phase retrieval task.}
\label{tab:transition-ablation}
\centering
\footnotesize
\setlength{\tabcolsep}{3.4pt}
\begin{tabular}{lcccc}
\toprule
Models & PSNR(dB) $\uparrow$ & SSIM $\uparrow$ & LPIPS $\downarrow$ & ArcFace $\uparrow$ \\
\midrule
Deterministic DDIM & 13.618 & 0.238 & 0.644 & 0.115 \\
Stochastic DDIM ($\eta=1$) & 16.495 & 0.394 & 0.545 & 0.276 \\
Point refresh ($M=1$) & 28.813 & 0.761 & 0.214 & 0.881 \\
DCTS ($M=8$) & \textbf{29.696} & \textbf{0.812} & \textbf{0.170} & \textbf{0.908} \\
\bottomrule
\end{tabular}
\end{table}

\section{Conclusion}

We proposed DCTS, which reformulates independent posterior refresh as direct conditional transport without explicit clean-space sampling and re-noising. The flow recovers the ideal refresh kernel under exact conditional-mean estimation and integration. A denoiser-compatible sufficient statistic and local Gaussian conditioning enable a training-free approximation. Experiments demonstrate competitive reconstruction quality with substantially reduced runtime. 

\bibliographystyle{IEEEbib}
\bibliography{strings,refs}

@inproceedings{ho2020ddpm,
  title     = {Denoising Diffusion Probabilistic Models},
  author    = {Ho, Jonathan and Jain, Ajay and Abbeel, Pieter},
  booktitle = {Advances in Neural Information Processing Systems},
  volume    = {33},
  pages     = {6840--6851},
  year      = {2020}
}

@inproceedings{song2021score,
  title     = {Score-Based Generative Modeling through Stochastic Differential Equations},
  author    = {Song, Yang and Sohl-Dickstein, Jascha and Kingma, Diederik P. and Kumar, Abhishek and Ermon, Stefano and Poole, Ben},
  booktitle = {International Conference on Learning Representations},
  year      = {2021}
}

@inproceedings{song2021ddim,
  title     = {Denoising Diffusion Implicit Models},
  author    = {Song, Jiaming and Meng, Chenlin and Ermon, Stefano},
  booktitle = {International Conference on Learning Representations},
  year      = {2021}
}

@inproceedings{karras2019stylegan,
  title     = {A Style-Based Generator Architecture for Generative Adversarial Networks},
  author    = {Karras, Tero and Laine, Samuli and Aila, Timo},
  booktitle = {Proceedings of the IEEE/CVF Conference on Computer Vision and Pattern Recognition},
  pages     = {4401--4410},
  year      = {2019}
}

@inproceedings{deng2019arcface,
  title     = {{ArcFace}: Additive Angular Margin Loss for Deep Face Recognition},
  author    = {Deng, Jiankang and Guo, Jia and Xue, Niannan and Zafeiriou, Stefanos},
  booktitle = {Proceedings of the IEEE/CVF Conference on Computer Vision and Pattern Recognition},
  pages     = {4690--4699},
  year      = {2019}
}

@inproceedings{chung2023dps,
  title     = {Diffusion Posterior Sampling for General Noisy Inverse Problems},
  author    = {Chung, Hyungjin and Kim, Jeongsol and McCann, Michael T. and Klasky, Marc L. and Ye, Jong Chul},
  booktitle = {International Conference on Learning Representations},
  year      = {2023}
}

@inproceedings{kawar2022ddrm,
  title     = {Denoising Diffusion Restoration Models},
  author    = {Kawar, Bahjat and Elad, Michael and Ermon, Stefano and Song, Jiaming},
  booktitle = {Advances in Neural Information Processing Systems},
  volume    = {35},
  pages     = {23593--23606},
  year      = {2022}
}

@inproceedings{zhu2023diffpir,
  title     = {Denoising Diffusion Models for Plug-and-Play Image Restoration},
  author    = {Zhu, Yuanzhi and Zhang, Kai and Liang, Jingyun and Cao, Jiezhang and Wen, Bihan and Timofte, Radu and Van Gool, Luc},
  booktitle = {Proceedings of the IEEE/CVF Conference on Computer Vision and Pattern Recognition Workshops},
  pages     = {1219--1229},
  year      = {2023}
}

@inproceedings{zhang2025daps,
  title     = {Improving Diffusion Inverse Problem Solving with Decoupled Noise Annealing},
  author    = {Zhang, Bingliang and Chu, Wenda and Berner, Julius and Meng, Chenlin and Anandkumar, Anima and Song, Yang},
  booktitle = {Proceedings of the IEEE/CVF Conference on Computer Vision and Pattern Recognition},
  pages     = {20895--20905},
  year      = {2025}
}

@inproceedings{janati2025mixture,
  title     = {A Mixture-Based Framework for Guiding Diffusion Models},
  author    = {Janati, Yazid and Moufad, Badr and Qassime, Mehdi Abou El and Oliviero Durmus, Alain and Moulines, Eric and Olsson, Jimmy},
  booktitle = {Proceedings of the 42nd International Conference on Machine Learning},
  series    = {{PMLR}},
  volume    = {267},
  pages     = {26830--26876},
  year      = {2025}
}

@article{efron2011tweedie,
  title   = {Tweedie's Formula and Selection Bias},
  author  = {Efron, Bradley},
  journal = {Journal of the American Statistical Association},
  volume  = {106},
  number  = {496},
  pages   = {1602--1614},
  year    = {2011}
}

@inproceedings{song2023pigdm,
  title     = {Pseudoinverse-Guided Diffusion Models for Inverse Problems},
  author    = {Song, Jiaming and Vahdat, Arash and Mardani, Morteza and Kautz, Jan},
  booktitle = {International Conference on Learning Representations},
  year      = {2023}
}

@article{yismaw2025codps,
  title   = {Gaussian Is All You Need: A Unified Framework for Solving Inverse Problems via Diffusion Posterior Sampling},
  author  = {Yismaw, Nebiyou and Kamilov, Ulugbek S. and Asif, M. Salman},
  journal = {IEEE Transactions on Computational Imaging},
  volume  = {11},
  pages   = {1020--1030},
  year    = {2025},
  doi     = {10.1109/TCI.2025.3594988}
}

@inproceedings{moufad2024mgps,
  title     = {Variational Diffusion Posterior Sampling with Midpoint Guidance},
  author    = {Moufad, Badr and Janati, Yazid and Bedin, Lisa and Durmus, Alain and Douc, Randal and Moulines, Eric and Olsson, Jimmy},
  booktitle = {International Conference on Learning Representations},
  year      = {2025}
}

@inproceedings{holderrieth2026glass,
  title     = {{GLASS} Flows: Transition Sampling for Alignment of Flow and Diffusion Models},
  author    = {Holderrieth, Peter and Singer, Uriel and Jaakkola, Tommi and Chen, Ricky T. Q. and Lipman, Yaron and Karrer, Brian},
  booktitle = {International Conference on Learning Representations},
  year      = {2026}
}

@inproceedings{lipman2023flow,
  title     = {Flow Matching for Generative Modeling},
  author    = {Lipman, Yaron and Chen, Ricky T. Q. and Ben-Hamu, Heli and Nickel, Maximilian and Le, Matt},
  booktitle = {International Conference on Learning Representations},
  year      = {2023}
}

@inproceedings{karras2022edm,
  title     = {Elucidating the Design Space of Diffusion-Based Generative Models},
  author    = {Karras, Tero and Aittala, Miika and Aila, Timo and Laine, Samuli},
  booktitle = {Advances in Neural Information Processing Systems},
  volume    = {35},
  pages     = {26565--26577},
  year      = {2022}
}

@article{boys2024tmpd,
  title   = {Tweedie Moment Projected Diffusions for Inverse Problems},
  author  = {Boys, Benjamin and Girolami, Mark and Pidstrigach, Jakiw and Reich, Sebastian and Mosca, Alan and Akyildiz, Omer Deniz},
  journal = {Transactions on Machine Learning Research},
  year    = {2024}
}

@inproceedings{peng2024covariance,
  title     = {Improving Diffusion Models for Inverse Problems Using Optimal Posterior Covariance},
  author    = {Peng, Xinyu and Zheng, Ziyang and Dai, Wenrui and Xiao, Nuoqian and Li, Chenglin and Zou, Junni and Xiong, Hongkai},
  booktitle = {Proceedings of the 41st International Conference on Machine Learning},
  series    = {{PMLR}},
  volume    = {235},
  pages     = {40347--40370},
  year      = {2024}
}

@inproceedings{wang2023ddnm,
  title     = {Zero-Shot Image Restoration Using Denoising Diffusion Null-Space Model},
  author    = {Wang, Yinhuai and Yu, Jiwen and Zhang, Jian},
  booktitle = {International Conference on Learning Representations},
  year      = {2023}
}

\end{document}